\documentclass[10pt,twocolumn,letterpaper]{article}

\usepackage{amsfonts}
\usepackage{amsmath}
\DeclareMathOperator{\Tr}{tr}
\usepackage{mathabx}					 	
\usepackage{placeins}                       
\usepackage{subcaption,caption}
\usepackage{epsfig}
\usepackage{booktabs}                       
\usepackage{tabularx}                       
\graphicspath{{Figures/}}                   
\usepackage{accents}                        
\usepackage{supertabular}
\usepackage{multirow}
\usepackage{longtable}
\usepackage{capt-of}                        
\usepackage{comment}                        
\usepackage{algorithm}                   
\usepackage{algpseudocode}
\usepackage[hyphens]{url}
\usepackage[colorlinks=true,urlcolor=blue]{hyperref} 

\usepackage{iccv}
\usepackage{times}
\usepackage{graphicx}
\usepackage{amssymb}
\usepackage{bbm}
\usepackage{mathrsfs}

\iccvfinalcopy 

\def\iccvPaperID{406} 

\begin{document}

\title{The GA-cal software for the automatic calibration of soil constitutive laws: a tutorial and a user manual}

\author{Francisco J. Mendez{$^{1,*}$},~ Miguel A. Mendez{$^{2}$},~Antonio Pasculli{$^{1}$}\\
{$^{1}$}{\small University G. D'Annunzio, Dept. of Engineering Geology (INGEO), Chieti-Pescara, Italy};\\ {$^{2}$} {\small von Karman Institute for Fluid Dynamics, EA Department, Sint-Genesius-Rode, Belgium}\\
{\small{{$^{*}$}Corresponding to: \tt{francisco.mendez@unich.it}}}}


\maketitle
\thispagestyle{empty}

\begin{abstract}
The calibration of an advanced constitutive law for soil is a challenging task. This work describes GA-cal, a Fortran software for automatically calibrating constitutive laws using Genetic Algorithms (GA) optimization. The proposed approach sets the calibration problem as a regression, and the GA optimization is used to adjust the model parameters so that a numerical model matches experimental data.
This document provides a user guide and a simple tutorial. We showcase GA-cal on the calibration of the Sand Hypoplastic law proposed by von Wolffersdorff, with the oedometer and triaxial drained test data. The implemented subroutines can be easily extended to solve other regression or optimization problems, including different tests and constitutive models. The source code and the presented tutorial are freely available at \url{https://github.com/FraJoMen/GA-cal}.
\end{abstract} 


\section{Introduction}

This work describes GA-cal, a Fortran software for automatically calibrating constitutive laws using Genetic Algorithms (GA) optimization. The calibration consist in identifying the model parameters associated with a specific soil. The proposed approach consists of three ingredients: (1) a solver that simulates the laboratory tests of interest for a given set of constitutive model's parameters, (2) a cost function measuring the performances of the model, i.e. how closely the model predicts the available data from the tests and (3) an optimization algorithm to look for the parameters that give the best prediction (that is a minimum of the cost function).

We here present the calibration of the Sand Hypoplastic law (SH) developed by von Wolffersdorff \cite{Wolffersdorff_A_hypoplastic_for_granular_material_with_a_predefined_limit_state_surface} using the oedometer (OE) and triaxial drained (TD) tests data. This approach was first presented in \cite{Mendez2021}, developed in a Python code and used to calibrate six of the eight coefficients in the SH model. The Fortran code presented in this document optimizes all eight parameters of the SH model and is 30 times faster than the previous Python code. Besides the speed up gained by the porting into Fortran, the new version benefits from a more consistent use of matrix multiplications and an improved approach for generating new individuals. Moreover, its current architecture offers specific subroutines for each of the three ingredients previously described. This allows for extending the proposed approach far beyond the analysed test case, including other constitutive laws and other experimental tests.

This document is organized as follows. In section 2, we briefly recall, in the interest of self-consistency, the SH constitute model currently available in GA-Cal. Section 3 reports on the proposed methodology while section 4 describes the GA-cal code. Section 5 presents a tutorial with an example calibration.

\section{The Sand Hypoplasticity (SH) model}\label{Par2}

The simulation of the OE and TD tests according to the SH model by \cite{Wolffersdorff_A_hypoplastic_for_granular_material_with_a_predefined_limit_state_surface} requires solving a system of nonlinear Ordinary Differential Equations (ODEs). These equations link the time derivative of the Cauchy effective stress $\dot{\mathbf{T}}$ and the void ratio $\dot{e}$, with the granulate stretching rate $\mathbf{D}$, the void ratio $e$ and the Cauchy effective stress $\mathbf{T}$.


In axisymmetric conditions such as those encountered in the OE and TD tests, denoting by $T_1$,$T_2$ and $D_1$,$D_2$ the principal (axial) and the second (radial) component of $\mathbf{T}$ and $\mathbf{D}$, 
\begin{equation}\label{eq:axisTD}
\mathbf{T}= \begin{bmatrix}T_1 & 0 &0 \\ 0 & T_2& 0 \\ 0&0&T_2 \end{bmatrix}  \quad  \mathbf{D}= \begin{bmatrix}D_1 & 0 &0 \\0 & D_2&0\\0&0&D_2 \end{bmatrix}\,,
\end{equation} the SH model gives: 

\begin{multline}
\begin{bmatrix}
       \dot{T}_1\\[0.3em]
       \dot{T}_2\\[0.3em]
       \dot{e}\\
      \end{bmatrix} \quad  = f_s \begin{bmatrix}
       L_{11} & L_{12}&0\\[0.3em]
       L_{21} & L_{22}&0\\[0.3em]
       0 & 0&1+e\end{bmatrix}\,\begin{bmatrix}
       D_1\\[0.3em]
       D_2\\[0.3em]
       D_1+2D_2\\
      \end{bmatrix}\,+
      \\  +f_s\,f_d \,
       \begin{bmatrix}
       N_1 \\[0.3em]
       N_2 \\[0.3em]
       0 \end{bmatrix} \sqrt{D_1^2+2D_2^2}\,,
\label{eq:sistema_SH}
\end{multline} where 

\begin{align}
&L_{11}	= \Tr(\mathbf{T})^2/\Tr(\mathbf{T^2})\cdot(1+a^2T_1^2/\Tr(\mathbf{T})^2)\\
&L_{12}	= 2a^2T_1T_2/\Tr(\mathbf{T}^2)\\
&L_{21}	= a^2T_1T_2/\Tr(\mathbf{T}^2)\\
&L_{22}	= 2a^2T_1T_2/\Tr(\mathbf{T}^2)\cdot(1+a^2T_2^2/tr(\mathbf{T})^2)\\
&N_{11}	= \Tr(\mathbf{T})/\Tr(\mathbf{T^2})\cdot a/3(5T_1-2T_2)\\
&N_{22}	= \Tr(\mathbf{T})/\Tr(\mathbf{T^2})\cdot a/3(4T_2-T_1) 
\end{align} and $\Tr$ denotes the trace of a tensor.

The coefficients $(a,f_s,f_d)$ have a semi-empirical interpretation and depend on the parameters of the model that needs to be tuned during the calibration. These reads:

\begin{align} \label{eq:a}
&a				=\frac{\sqrt{3}(3-\sin \varphi_c)}{2\sqrt{2}\sin \varphi_c }\,;
\end{align}

 \begin{equation}
f_s=\cfrac{\cfrac{h_s}{n}\cfrac{1+e_i}{e_i}\Biggr(\cfrac{e_{i}}{e}\Biggl)^\beta \,\Biggr(-\cfrac{\Tr(\mathbf{T})}{h_s}\Biggl)^{1-n}}{3+a^2-a\,\sqrt{3}\, \Biggr( \cfrac{e_{i0}-e_{d0}}{e_{c0}-e_{d0}}\Biggl)^\alpha}\,;
\label{eq:f_s}
\end{equation}

\begin{equation}
f_d=\Biggr(\frac{e-e_d}{e_c-e_d}\Biggl)^\alpha\,.
\label{eq:f_d}
\end{equation}

The previous equations  depend  on the    maximal ($e_d$), minimal ($e_i$) and critical ($e_c$)  void   ratio. These are linked, according to 
\cite{Bauer_Calibration_of_a_comprehensive_hypoplastic_model_for_granular_materials}, by the  system:
 
\begin{equation}\label{eq:ei_ed_c_ed}
\frac{e_i}{e_{i0}}=\frac{e_d}{e_{d0}}=\frac{e_c}{e_{c0}}=\exp\Biggr[-\Biggr(\frac{-\Tr(\mathbf{T})}{h_s}\Biggl)^n\Biggl]\,.
\end{equation}

For a given mean pressure, among the possible void ratios $e_d<e<e_i$, we can identify regions of dilative (for $e_d<e<e_c$) and contractive (for $e_c<e<e_i$) behaviour. The ratios $\lambda_d=e_{d0}/e_{c0}$ and $\lambda_i=e_{i0}/e_{c0}$ govern the amplitude of the domains of dilatant or contractive behaviour, while $e_{c0}$ defines the critical state in terms of void ratio. 

The set of equations \eqref{eq:a}-\eqref{eq:ei_ed_c_ed} thus includes eight parameters, i.e. $\{  \varphi_c , h_s , n ,  e_{c0} , \alpha ,  \beta, e_{d0},  e_{i0}  \}\,$. The reader is referred to \cite{Herle_Gudehus_Determination_of_parameters_of_a_hypoplastic_constitutive_model_from_properties_of_grain_assemblies,Masin_The_influence_of_experimental_and_sampling_uncertainties_on_the_probability_of_unsatisfactory_performance_in_geotechnical_applications,Mendez2021} for more details on these parameters.

The GA used in the calibration involves randomly generating different possible combinations of parameters (later referred to as candidate solutions). To be valid, the inequality $ e_ {i0}< e_ {c0} <e_ {d0} $ must be respected at all times. To enforce this, we compute $e_{d0}$ and $e_{i0}$ using the ratio 
previously defined $\lambda_d<1$ and $\lambda_i>1$. Experimentally it has been observed that $\lambda_d=0.52 \div 0.65 $ and $\lambda_i\cong 1,2$ \cite{Herle_Gudehus_Determination_of_parameters_of_a_hypoplastic_constitutive_model_from_properties_of_grain_assemblies}. Therefore, the set of parameters calibrated by the GA is:

\begin{equation}
  \mathbf{P}=\{ \varphi_c , h_s , n , e_{c0} , \alpha ,  \beta, \lambda_d , \lambda_i  \}\,.  
\end{equation}

\section{Integration Procedure}\label{TrepUno}
The dynamical system \eqref{eq:sistema_SH} is integrated using the simple explicit Euler scheme. This allows for an easy check of the solution admissibility ($\Tr(\mathbf{T})<0$ and $e_d\leq e \leq e_i$) at every time step. Defining $\mathbf{X}^{k}=[T_1^{k}, T_2^{k}, e^{k}]^T$ the state vector of the ODE system, the time integration scheme reads:
\begin{equation}
\label{eq:Euler_int}
 \mathbf{X}^{k+1}=\mathbf{H}(\mathbf{X}^{k},\mathbf{P})\,\Delta t +\mathbf{X}^{k}\,,
\end{equation} where $\mathbf{H}$ is the nonlinear function that returns the right term of the \eqref{eq:sistema_SH}.

The SH constitutive law is a state-dependent model homogeneous of the first order in $\mathbf{D}$ \cite{Niemunis_Extended_hypoplastic_models_for_soils}. So it is possible to select an arbitrary reference $D_1=-1$ and compute the integration time $t_f$ from the maximal deformation obtained at the end of each test. For the OE test, the integration time is 
\begin{equation}
\label{eq16}
    t_f =  -\ln \biggr( 1- \frac{e_{0}-e_{fin}}{e_{0}+1}\biggl)\,,
\end{equation} while for the TD test it is 
\begin{equation}
\label{eq17}
  t_f = \varepsilon_{fin}  \,.
\end{equation}
The parameters $e_0,\,e_{fin}$ in \eqref{eq16} are the initial void ratio and the final void ratio, namely at the end of the compression part in the OE test. The parameter $\varepsilon_{fin}$ in \eqref{eq17} is the maximum axial deformation in the TD test. Once the integration time $[0, \,t_f]$ is defined, the time step is computed as $\Delta t=t_f / n_{Step}$, where the number of time steps can be set to $n_{Step}=100$. The function $\mathbf{H}(\mathbf{X},\mathbf{P})$ in \eqref{eq:Euler_int} differs in the two tests, as detailed in \cite{Mendez2021}.

In the Ga-cal software, the function $\mathbf{H} $ in \eqref{eq:sistema_SH} for the OE and TD tests are computed by the subroutines 
\texttt{dydtOE} and \texttt{dydtTD} respectively.

\section{Calibration Methodology}\label{Quattro}
The proposed calibration procedure combines a stochastic optimizer with a numerical solver of Ordinary Differential Equations (ODEs). For every set of parameters $\mathbf{P}$, the ODE solver integrates the SH model described in the previous section to compute the soil response to a set of tests while the optimizer compares the resulting curves with the experimental data points and updates the parameters $\mathbf{P}$ 
until a maximum number of iteration is reached. 

The following subsections describe the optimization algorithm implemented in GA-cal, focusing on the cost-function definition in \ref{QuattropUno} and the GA strategy in \ref{QuattropDue}.

\subsection{Cost-function Definition}\label{QuattropUno}
The cost-function driving the optimization of the parameters $\mathbf{P}$ is built to account for the discrepancy between the numerical predictions and the set of measurements \cite{https://doi.org/10.1002/nag.2714}. 

For a given set of parameters $\mathbf{P}$, the subroutines \texttt{intE\_OE} and \texttt{intE\_TD} are used to simulate the soil response to the OE and TD tests respectively. For each of these tests, these subroutines provide the evolution of the state vector $\mathbf{X}(t_k)$ in the dynamical system \eqref{eq:sistema_SH}. The results are then manipulated prior to the comparison with experimental data. Specifically, the results of the OE tests are provided in the plane ($\sigma_v$,\,$e$), where $\sigma_v=-T_1$ is the vertical compression pressure; hence we are interested in the temporal evolution of the first and third components of the vector $\mathbf {X}(t_k)$. The results of the TD test are given in the triaxial deviatoric diagram ($\varepsilon_a,\,q$) and triaxial volumetric diagram ($\varepsilon_a,\,\varepsilon_v$), thus the temporal evolution of the variables of interest are:

\begin{itemize}
    \item axial deformation
\begin{equation}\label{eq:varepsilon_a}
    \varepsilon_a(t)=-\int_0^{t} D_1 \,dt\,;
\end{equation} 
 
 \item volume change $\varepsilon_v$

\begin{equation}\label{eq:varepsilon_v}
    \varepsilon_v(t)=-\int_0^{t} \Tr(\mathbf{D}) \,dt=\frac{e(t)-e_{0}}{1+e_{0}}\,;
\end{equation}

\item deviatoric stress 

\begin{equation}\label{eq:q}
    q(t)=T_2(t)-T_1(t)\,.
\end{equation} 

\end{itemize} These three variables are computed using the subroutine \texttt{elb\_TD}. 

These results are projected in dimensionless plane before evaluating the model performances, to cope with the largely different numerical values taken by the variables \cite{Mendez2021}.   The dimensionless variables are computed as follows 

\begin{alignat}{1}
&\widehat{\varepsilon}_a =\varepsilon_a/\varepsilon_{fin}\,;\\
&\widehat{\varepsilon}_v =\varepsilon_v/max(|\varepsilon_v|)\,;\\
&\widehat{q}\,\, = q/max(q)\,;\\
&\widehat{T}_1 = -T_1/max(|T_1|)\,;\\
&\widehat{\varepsilon}_E =\ln\Bigr(1-\frac{e_0-e}{e_0+1}\Bigl)/\ln\Bigr(1-\frac{e_0-e_{fin}}{e_0+1}\Bigl) \,.
\end{alignat} 

 The data from the OE tests are mapped to $\Pi_1=(\widehat{\varepsilon}_E,\widehat{T}_1)$, while the data from TD tests are mapped to $\Pi_2=(\widehat{\varepsilon}_a,\widehat{q})$ and $\Pi_3=(\widehat{\varepsilon}_a,\widehat{\varepsilon}_v )$. As a result of the scaling, the experimental data for each tests start at ($0,0$) and end at ($1,1$).  The scaling is performed in the subroutines \texttt{Cost\_OE} and \texttt{Cost\_TD}.
 
\begin{figure}[hbt]
  \centering
    \includegraphics[width=0.43\textwidth]{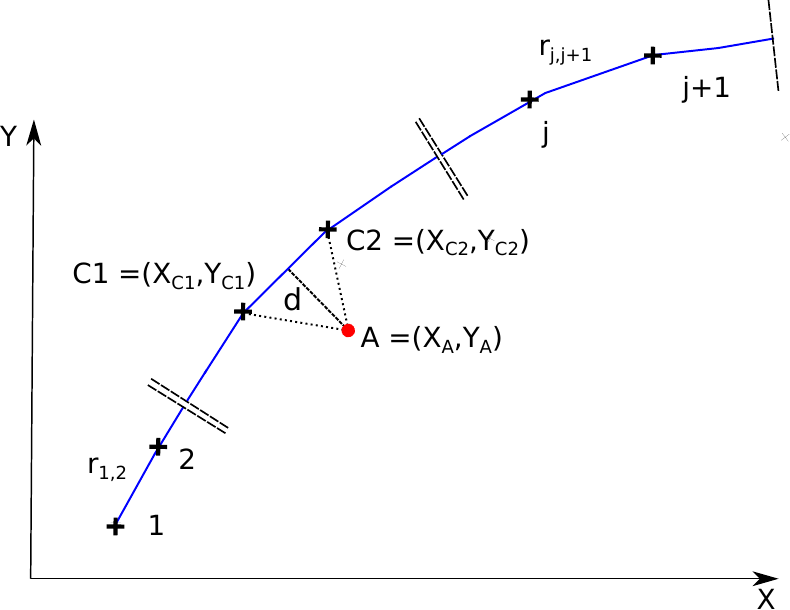}
\caption{Definition of discrete  Fr\'{e}chet distance $d$ between a points $A(X_A,Y_A)$ and a numerical curve $r_{j,j+1}$.}
\label{Example_NORMA}
\end{figure}

 The discrepancy between model predictions and data is measured in terms of mean of the Discrete Fr\'{e}chet Distance (DFD). This has the main merit of being invariant to the axis orientation \cite{Jekel2018}. The DFD is briefly introduced in Figure \ref{Example_NORMA}. Given the set of predictions of the SH model $(X_j,Y_j)$ at times $t_j$, we denote as $r_{j,{j+1}}$ the segment connecting the points $j$ and $j+1$,
 so that the set of segments define a piece-wise linear approximation of the numerical response curve. Then, given the experimental point $A = (X_A, Y_A) $, the DFD is the distance between the point $A$ and the closest segment, i.e. the one connecting the closest points (C1 and C2 in Figure \ref{Example_NORMA}).

Denoting as $\mathbf{d}^{i,j}_{\mathcal{F}}$ the vector collecting the DFD of all points on the plane $\Pi_i$, with $i=1,2,3$, for the tests $j$, we define a performance metrics as 

\begin{equation}\label{eq:delta_i}
 \delta_i=\frac{1}{n_j}\sum^{n_{i}}_{j}||\mathbf{d}^{i,j}_{\mathcal{F}}||_2  \end{equation} where $n_{i}$ is the number of tests in the i-th plane, $n_j$ is the number of points for each test and $||\,.\,||_2$ is the $l_2$ norm of a vector. This metrics is a function of the parameters $\mathbf{P}$ used for the model prediction. Therefore, we define the global performance of these parameters as 

\begin{equation}
\label{eq:Sco}
Cost\,(\mathbf{P})= w_1 \,\delta_1(\mathbf{P}) + w_2 \,\delta_2(\mathbf{P})+ w_3 \,\delta_3(\mathbf{P}) 
\end{equation} where $w_i$, with $i=1,2,3$, weights the relative importance of each plane $\Pi_1,\Pi_2, \Pi_3$. The best set of parameters is the one that minimizes the cost function \eqref{eq:Sco}.

It is worth noticing that the cost function in \eqref{eq:Sco} can be easily extended to include results from other tests. Moreover, one could treat the performances $\delta_i$ in each i-th plane as an independent cost function and thus move from the current single-objective optimization to a multi-objective optimization \cite{Hiroyasu2005}. This extension is left to future work.

In Ga-cal, the cost function \eqref{eq:Sco} is computed by the subroutine \texttt{Cost\_Pop}, while the performances in the three planes are computed by the subroutines \texttt{Cost\_OE} and \texttt{Cost\_TD}.


\subsection{Genetic Algorithm (GA) Optimizer}\label{QuattropDue}

The GA is a population-based stochastic optimizer which operates on a set (population) of possible candidate solution (here set of parameters $\mathbf{P}$).
These are evaluated and updated using a set of statistical operators that mimic the Darwinian theory of survival of the fittest \cite{10.7551/mitpress/1090.001.0001}.

The optimization thus consists of three steps: (1) initialize, (2) evaluate and (3) update the population. The steps (2) and (3) are repeated in loop until maximum number of iterations $N_{ITER}$ is reached. The best individual $\mathbf{P^*}$ is taken from the last update of the population.

The initialization consists in randomly creating a set of candidate solutions. The evaluation consists in running the SH model for each of these and compute the associated cost according to the previous section. The population update uses a set of operations to increase the likelihood of identifying a minimum with respect to a simple random search. Both the steps (1) and (2) involves the generation of random numbers. This is described in \ref{4p2p1}. The initialisation, evaluation and the population update are briefly described in \ref{4p2p2}, \ref{4p2p3} and \ref{4p2p4}. The subroutines linked to each of these steps are documented in Section \ref{Cinque}.

\subsubsection{Random numbers generation}
\label{4p2p1}
Following the continuous GA in \cite{Haupt2003}, the main mechanism for random number generation is the sampling of a uniform distribution in the range $ [0,1) $. This was carried out using the \texttt{random\_number} subroutine of the GNU Fortran compiler. 

The uniform distribution can then be used to generate other distributions such as the Gaussian and log-normal distributions \cite{PASCULLI2018370,doi:10.1063/5.0027121}. In this work, we use it to generate the triangular distribution because it allows for a good compromise between `\textit{exploration}' (i.e. the tendency to search in unexplored area) and `\textit{exploitation}' (i.e. the tendency to focus on the known minima) \cite{Haupt2003}.

\begin{figure}[hbt]
  \centering
    \includegraphics[width=0.40\textwidth]{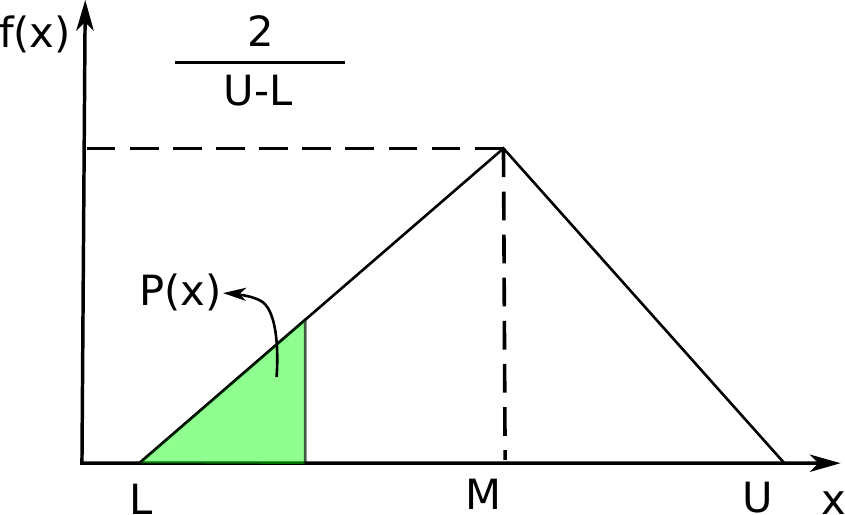}
\caption{Probability density function $f(x)$ for a triangular distribution.}
\label{fig:dist_triang}
\end{figure}

The triangular distribution, illustrated in Figure \ref{fig:dist_triang} is univocally defined by three numbers, i.e. the lower and upper bounds ($L$ and $U$) and the mode ($M$). Given a number $P\in [0,1)$, sampled from a uniform distribution, the triangularly distributed random variable $x$ reads:

\begin{equation}
x = \begin{cases}
L+\sqrt{P\,(U-L)\,(M-L)} &\text{$0<P<F$}\\
L-\sqrt{(1-P)\,(U-L)\,(U-M)} &\text{$F<P<1$}
\end{cases}
\end{equation} where
\begin{equation}
    F=\frac{M-L}{U-L}
\end{equation}

Ga-cal uses both the uniform and triangular distributions as detailed in the next subsections.

\subsubsection{Population Inizialization}\label{4p2p2}

GA-cal encodes a population of candidates (individuals) in the matrix $\mathbf{Pop}$ of size ($N_i$,$P_{dim}$), where $N_i$ is the number of individuals, and $P_{dim}$ is the dimension of the optimization problem (in this case $P_{dim}=8$ is the size of the parameters $\mathbf{P}$). Thus each individual is stored as a row of $\mathbf{Pop}$.

The population is initialized randomly within the search space bounded by vectors containing the lowest and the largest possible values of each parameter $\mathbf{P}_{min}$ and $\mathbf{P}_{max}$. This operation is made using a matrix of random numbers $\mathbf{\Theta}\in\mathbb{R}^{N_i,P_{dim}}$, uniformly distributed in the range $[0,1)$:

\begin{equation}
    \mathbf{Pop}(i,j)=\mathbf{P}_{min}(j)+\mathbf{\Theta}(i,j)\cdot(\mathbf{P}_{max}(j)-\mathbf{P}_{min}(j))    
\end{equation} where $i=1,2,...,N_{i}$ and $j=1,2,...,P_{dim}$.

\subsubsection{Population Evaluation}\label{4p2p3}
The evaluation procedure consists of two steps. The first step computes the cost-function \eqref{eq:Sco} for each set of parameters, i.e., for each individual. The second step ranks the individuals according to the cost function, used as a fitness measurement, from the best (low cost) to the worst (high cost) candidate solution. 
The evaluation returns the vector $\mathbf{Class}$, containing the index of first-placed individuals.

\subsubsection{Population update}\label{4p2p4}

The statistical operations involved in the population update are referred to as \emph{elitism}, \emph{mutation}, \emph{selection} and \emph{cross-over}. These are briefly described in the following.

\begin{itemize}

\item \emph{Elitism} consists in advancing some of the best individuals to the next generation. The elite individuals are taken from $\mathbf{Pop}$ using the pointer from the first element of the $\mathbf{Class}$ vector. Denoting with $N_{eli}$ the number of elite individuals passed to the next generation we have:

\begin{equation}\label{eq:Peli}
    \mathbf{P}_{eli}=\mathbf{Pop}(\mathbf{Class}(i)) 
\end{equation} where $i=1,2,...,N_{eli}$. 

\item \emph{Mutation} is one of the operations related to the exploration of the parameter space: a randomly chosen portion of the population ($N_{mut}$) at each iteration is randomly mutated according to two mechanisms: (1) recombination ($N_{com}$) and (2) re-initialization ($N_{ran}$). 

The recombination consists in combining the entries of the most successful candidates (here referred to as `genes'). The selection of the involved entries is carried out using the matrix $\mathbf{\Psi}$, which stores in each column a sample from a triangular distribution with $L=M=1$ and $U=N_{i}$ (rounded to the nearest integer). This the portion of recombined individuals reads

\begin{equation}
    \mathbf{P}_{com}(i,j)=\mathbf{Pop}(\mathbf{Class}(\mathbf{\Psi}(i,j)),j)\,    
\end{equation}where $i=1,2,...N_{com}$ and $j=1,2,..,P_{dim}$. 

The re-initialization consists in initializing $ N_{ran} $ individuals from scratch, using the same approach used in the initialization of the population.

To balance exploration and exploitation as the optimization progresses, the number of individuals undergoing mutation is progressively reduced. This procedure is akin to the annealing scheduling in simulated anealing. The reduction is defined as

\begin{equation}
   \mu = \mu_{in}  \biggl(\frac{\mu_{en}}{\mu_{in}}\biggr)^{\frac{ITER - 1}{N_{ITER}-1}}\,. 
\end{equation} with $ITER$ the iteration counter, $\mu_{in}$ and $\mu_{ed}$ the initial and final ratio of mutation. The evolving number of mutated individuals is thus updated as $N_{mut} \leftarrow \mu N_i $, rounded to the nearest digit. 
Concerning the partitioning into recombined and re-initialized individuals, $N_{com}$ is linearly increased with the iteration to further promote exploitation:

\begin{equation}
   N_{com}= N_{mut}\,\frac{ITER}{N_{ITER}}\,, 
\end{equation} while the number of reinitialised individuals is simply $N_{ran}=N_{mut}-N_{com}$.

\item \emph{selection and cross-over} are used to compute the remaining $N_{mat}=N_{i}-N_{eli}-N_{mut}$ individuals. 
The \emph{selection} defines which of the individuals mate while \emph{cross-over} defines how the mating produces new individuals from the selected ones.

The Ga-cal follows the weighting approach in \cite{Haupt2003}. Two random vectors  $\mathbf{X}$ and $\mathbf{Y}$, herein denoted as parents, are sampled from a triangular distribution with $L=M=1$ and $U=N_f=n_f\,N_{i}$ rounded to the closest integer, where $n_f$ is the fraction of individuals allowed to mate.
Being the population sorted by their fitness, this gives the best individuals a higher chance of being selected. After selection, the \emph{cross-over} is performed by randomly blending genes between the parents as 

\begin{multline}
\mathbf{P}_{mat}(i,j)= \mathbf{Pop}(\mathbf{Class}(X(i),j))\cdot \mathbf{\Theta}(i,j)
 \,+\\ +
 \mathbf{Pop}(\mathbf{Class}(Y(i),j))\cdot (1-\mathbf{\Theta}(i,j))\,,
\label{eq:P_mat}
\end{multline} where $i=1,2,...,N_{mat}$, $j=1,2,...,P_{dim}$ and  $\mathbf{\Theta}$ is matrix $(N_{mat},P_{dim})$ of random numbers uniformly distributed in the range $[0,1)$. 

\end{itemize}

After the three aforementioned operations are carried out, at each iteration the updated population is 

\begin{equation}
    \mathbf{Pop}_{New}=\mathbf{P}_{eli} \cup \mathbf{P}_{com} \cup \mathbf{P}_{ran} \cup \mathbf{P}_{mat} \,.
\end{equation}

\section{ The GA-cal code}\label{Cinque}
The GA-cal program is written in Fortran and is structured in a main file GA-cal\_main.f90 and four modules: 

\begin{itemize}
    \item Read\_input
    \item Create\_Pop
    \item Evaluate\_Pop
    \item Write\_out 
\end{itemize}

The code can be compiled using \emph{gfortran}\footnote{\url{https://gcc.gnu.org/fortran/}}. The \emph{Code::Blocks}\footnote{\url{http://www.codeblocks.org }} project attached to the source code 
allows to set up the compilation with all modules.
Once the compilation is completed, the executable should be copied in a folder containing a subfolder named \texttt{data} where all the inputs files, described in the following, should be collected. 

Running the executable produces a prompt message requesting the path and the name of an input file `\textit{data/file name.txt}'. Afterwards, the results obtained for the different iterations are displayed on the screen and all outputs are stored in the directory from which the executable is launched. The plots in the post-processing can be created with Python using a script like py\_OUT.py, attached in the program documentation.

The program's input and output are briefly described in the \ref{5p1} and \ref{5p2}, while the subsection from \ref{5p3} to \ref{5p6} provides a brief description of the GA-cal modules. The reader is referred to the comments in the code for more details.

\subsection{Input files}\label{5p1}
The input data required in the subfolder \textit{data} consists of seven text files.

\begin{description}
    \item [Input.txt] containing (1) the number of OE tests and TD tests, (2) the number of steps $n_{Step}$ in the time integration of the constitutive law \eqref{eq:Euler_int}, (3) the weights for the equation \eqref{eq:Sco} and (4) the names of the files that the program will read. 

The structure of this file is:
\end{description}
    \includegraphics[width=0.45\textwidth]{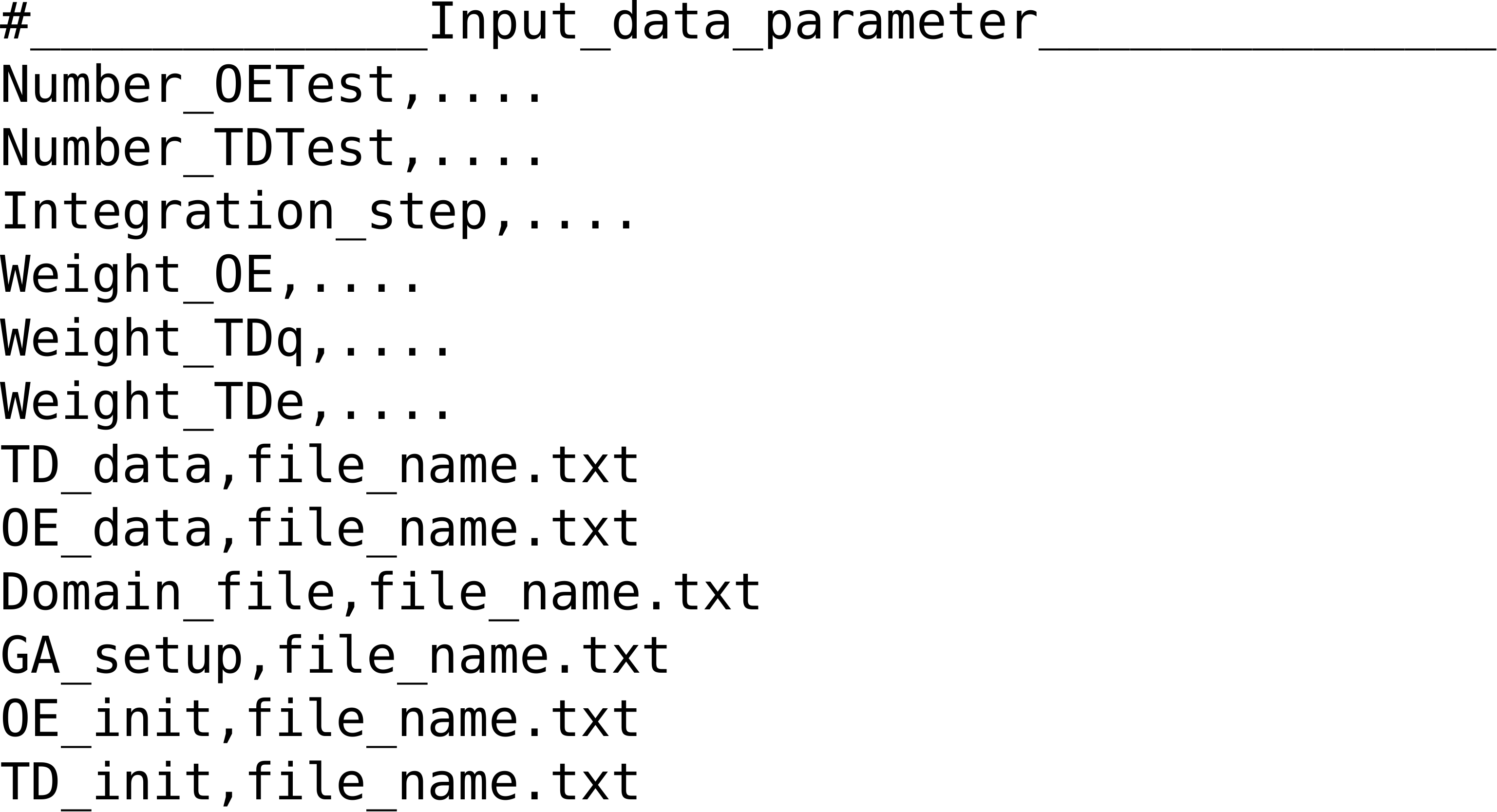}

\begin{description}   
    \item [Domain\_file.txt] contains the search range for each parameter, that is $\mathbf{P}_{min}$, $\mathbf{P}_{max} $. The number of parameters must be consistent with the $P_{dim}$ in the \texttt{GA\_setup.txt} file.
    
    The structure of this file is:
\end{description}

\includegraphics[width=0.45\textwidth]{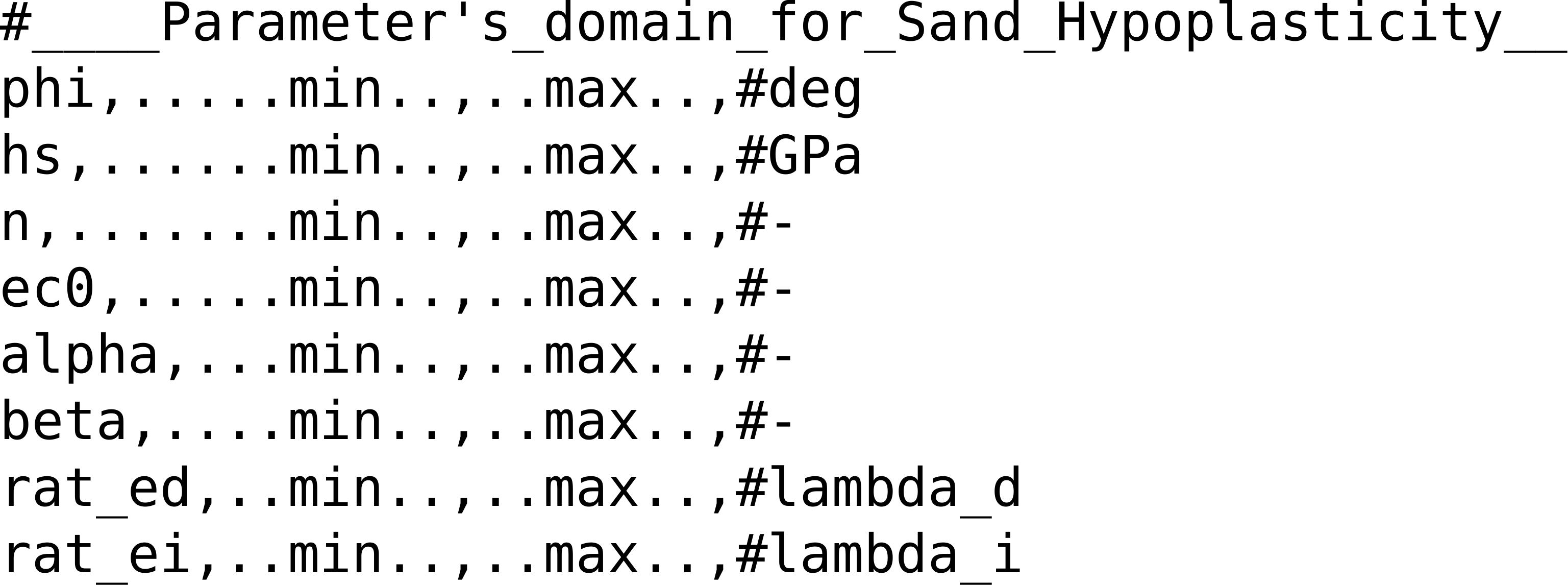}

\begin{description}
    \item [GA\_setup.txt] contains the parameters that controlling the optimizer. These are (1) the numbers of the individual $N_i$, (2) the problem dimensions $P_{dim}$ ($=8$ for the SH model), the number of iterations $N_{ITER}$ (3) the number of elite $N_{eli}$, (4) the fraction of individuals that are allowed to mate $n_f$ and (5) the final and initial ratio of mutation $\mu_{ed}$ and $\mu_{in}$.
    
    The structure of this file is:  
\end{description}

\includegraphics[width=0.45\textwidth]{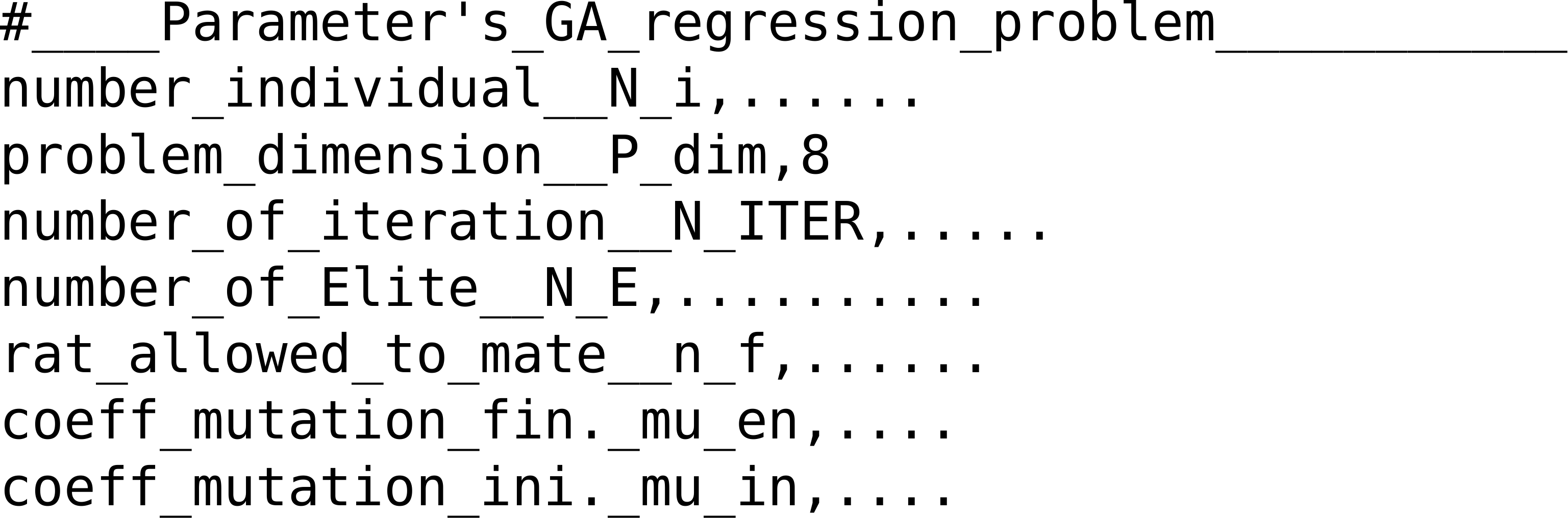}

\begin{description}       
    \item [OE\_init.txt] contains the initial conditions of each OE test, in the form of a vector. The length of this vector must be consistent with the number of data reported in the \texttt{Input.txt} file. The required initial conditions are (1) the principal (axial) and the second (radial) component of the Cauchy effective stress $T_1$ and $T_2$ (positive if the sample is compressed), and (2) the void ratio $e$.
    
    The structure of this file is:  
    \end{description}

\includegraphics[width=0.45\textwidth]{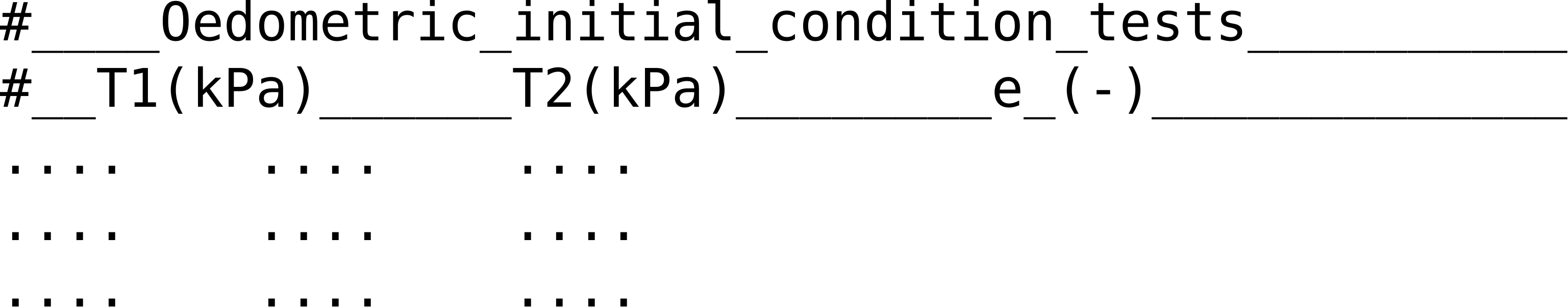} 

\begin{description}
    \item [TD\_init.txt] contains the initial conditions of each TD test. As for the previous file, these must be compatible with the number of data indicated in the \texttt{Input.txt} file. The initial conditions to prescribe are organised as fin the previous file.
    
    The structure of this file is thus identical to the previous one:
\end{description}

\includegraphics[width=0.45\textwidth]{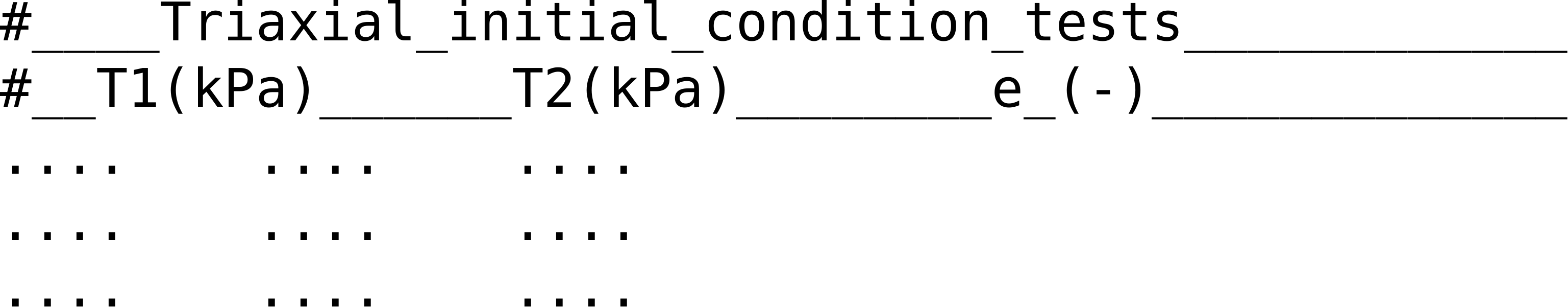} 

\begin{description}
    \item [OE\_data.txt] contains the experimental data collected from the OE tests.  Specifically, proceeding in column order, these are (1) the void ratio $e$ (2) the vertical stress $\sigma_v=-T_1$ (3) the identification number for the test $id$, ordered in ascending order.
    
    The structure of this file is:  
\end{description}
    
\includegraphics[width=0.45\textwidth]{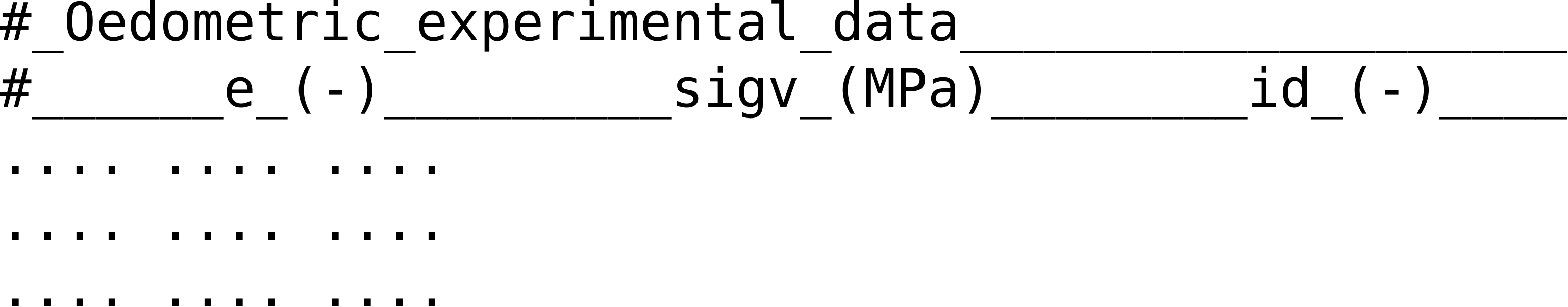}

\begin{description}
    \item [TD\_data.txt] contains the experimental data from the TD tests. In column order, these are (1) the axial deformation $\varepsilon_a$ (see  \eqref{eq:varepsilon_a}) (positive for the compression), (2) the volumetric deformation $\varepsilon_v$ (see  \eqref{eq:varepsilon_v}) (positive for the compression), (3) the deviatoric stress $q$ (see  \eqref{eq:q}) and (4) the identification number for the test $id$, ordered in ascending order.
    
    The structure of this file is: 
\end{description}
\includegraphics[width=0.45\textwidth]{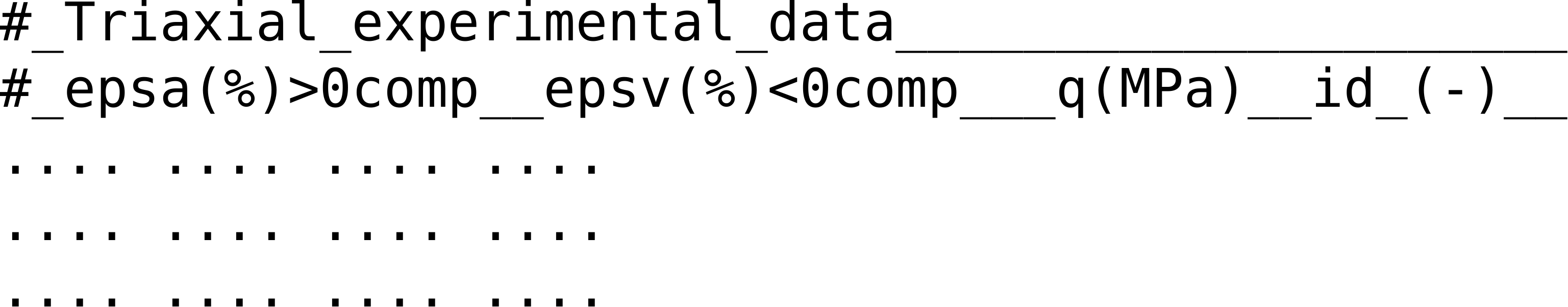}

\subsection{Output files}\label{5p2}
At the end of each execution, the program writes the following files:
\begin{description}
    \item [log\_Pop.txt]is a text file containing all combinations of parameters evaluated in the last run. This file is updated at each execution of GA-Cal.
    \item [best\_fit.txt] is a text file containing the optimal combination of $\mathbf{P}$ after each run.
    
    \item [data\_structure.dat] is a binary file with the information required to encode the following binary file in output.
    
    \item [X\_OE.dat] is a binary file collecting the coordinates of the experimental points of all the processed OE tests (in the plane $\sigma_v,e$).
    
    \item [X\_TD.dat] is a binary file collecting the coordinates of the experimental points in the plan ($\varepsilon_a, q $) and ($\varepsilon_a, \varepsilon_v $).
    
    \item [HX\_TD.dat] is a binary file collecting the response curves obtained with the last $ \mathbf{Pop}$ individuals in the planes ($\varepsilon_a, q$) and ($\varepsilon_a, \varepsilon_v$).
    
    \item [HX\_OE.dat] is a binary file collecting the response curves obtained with the last $ \mathbf{Pop} $ individuals in the odometer plane ($\sigma_v,\,e$).
\end{description}


\subsection{Read\_input}\label{5p3}
Read\_input is the first module executed by the program. This reads the input files in this module, and defines all common variables. The code requires the input file name specification, from which the program obtains all the information necessary for the calibration.  

\subsection{Create\_Pop}\label{5p4}
The Create\_Pop module contains two subroutines; one for creating the initial population and one for updating it. 

\begin{itemize}
\item \texttt{Initial\_Pop} computes the initial population by continuous uniform distribution between the Lower and Upper values of each model parameter defined in the input.

\item \texttt{Update\_Pop} uses elitism, mutation, selection and cross-over to generate a new population at each iteration.

\end{itemize}

\subsection{Evaluate\_Pop}\label{5p5}
This module contains eight subroutines for evaluating the cost-function and one for reordering the population and getting the vector $\mathbf{Class}$.

\begin{itemize}
\item \texttt{COST\_OE} compute the cost-function contribution in the oedometric plane.

\item \texttt{COST\_TD} compute the cost-function contribution in the triaxial planes.

\item \texttt{COST\_Pop} combines the computed errors with the subroutines \texttt{COST\_OE} and \texttt{COST\_TD} for the definition of the $Cost$ assigned to each individual of the population. The higher it is, the greater the deviation of the individual from the experimental data.

\item \texttt{dydtEDO} compute the time derivative of the ODE system of oedometric test. 
\item \texttt{dydtTxD}. compute the time derivative of the ODE system of triaxial drained test. 

\item \texttt{intE\_OE} perform the explicit Euler integration to simulate oedometer test using dydtEDO subroutine. 
\item \texttt{intE\_TD} perform the explicit Euler integration to simulate triaxial drained test using dydtTxD subroutine.
\item \texttt{elab\_TD} perform the elaboration of the triaxial drained test response and from ($T_1$ [MPa],$T_2$ [MPa], $e$[-]) compute ($q$ [MPa], $\varepsilon_a$ [\%], $\varepsilon_v$[ \%])

\item \texttt{merge\_argsort} returns a vector of indices that sorts the input vector in ascending order.

\end{itemize}

In addition to some specific functions used to integrate the constitutive model selected, this module contains the function \texttt{Frechet\_dist} to compute the averaged Fréchet distances between the experimental points and the response curves numerically obtained.

\subsection{Write\_out}\label{5p6}
 The Write\_out module writes all the output files described in the subsection \ref{5p2} expect for the files \textit{log\_PoP.txt}, written by the subroutines Create\_Pop, and \textit{best\_fit.txt} file, written by the GA-cal\_main. 
 
\section{An Example of Calibration}\label{Sei}
We here illustrate an example of GA-cal calibration of the SH model using the experimental data obtained from the two OE and the three TD tests reported in \cite{Wolffersdorff_A_hypoplastic_for_granular_material_with_a_predefined_limit_state_surface}. The selected experimental data, with all the input files used, can be found in the example folder in the repository of the source code.

\begin{figure}[htbp!]
  \centering
  \subcaptionbox{\label{fig:CLI_1}}[1\linewidth][c]{%
    \fbox{\includegraphics[width=0.45\textwidth]{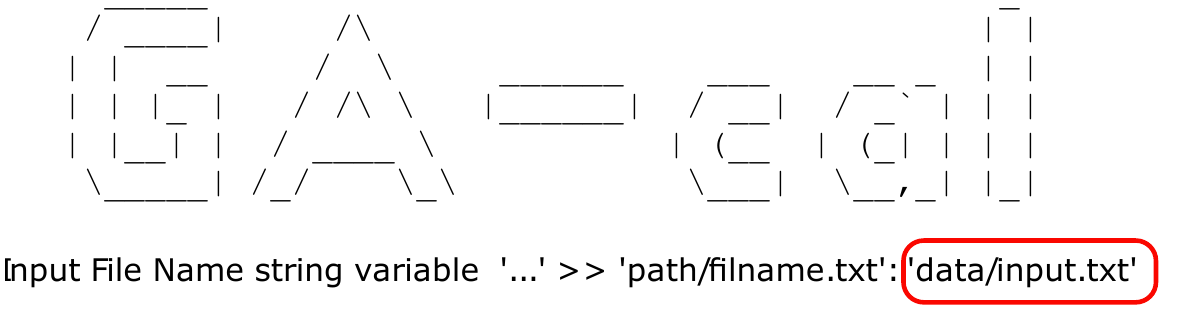} }} \\
  \subcaptionbox{  \label{fig:CLI_2}}[1\linewidth][c]{%
    \fbox{\includegraphics[width=0.45\textwidth]{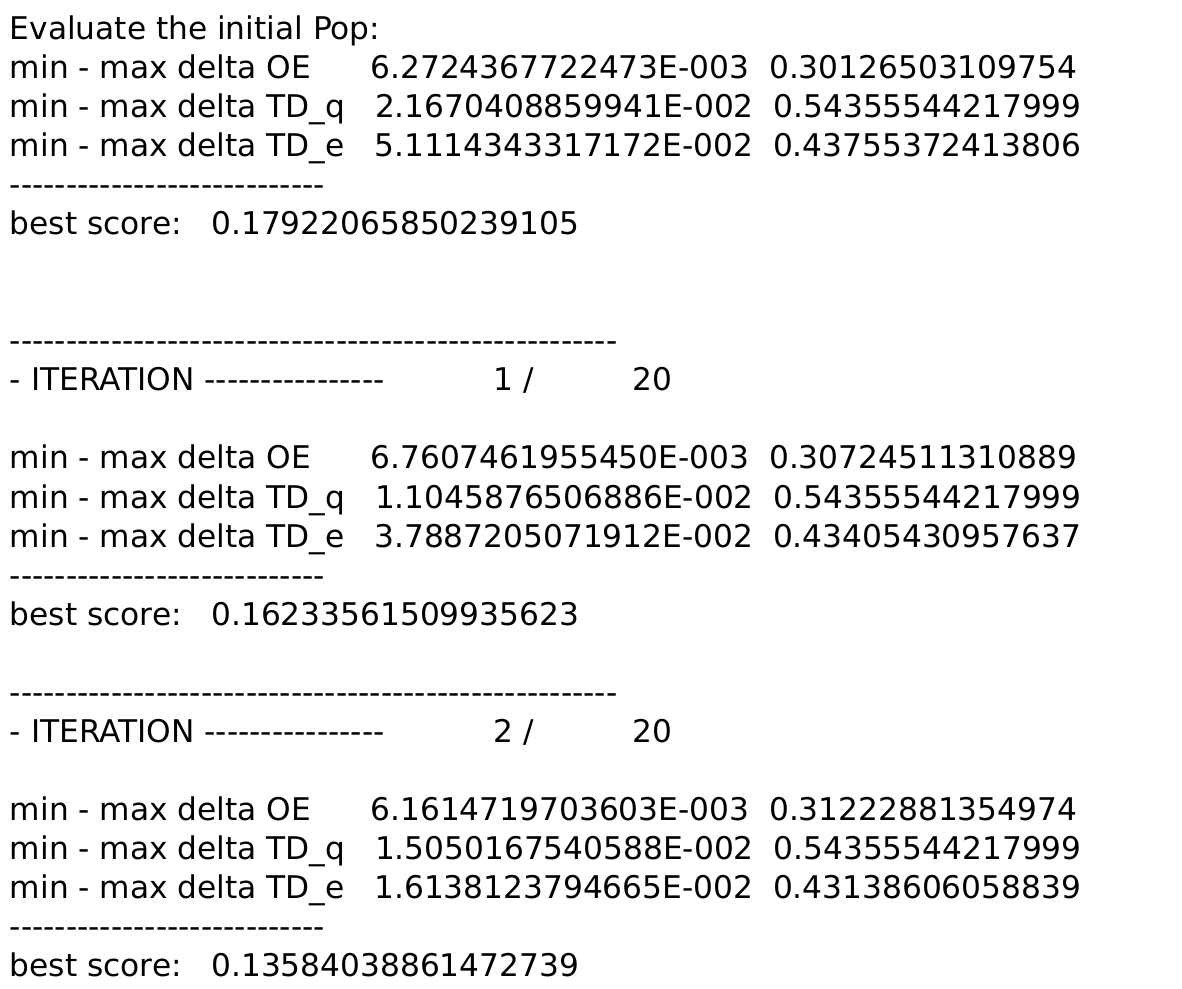} }} \\
  \subcaptionbox{ \label{fig:CLI_3}}[1\linewidth][c]{%
    \fbox{\includegraphics[width=0.45\textwidth]{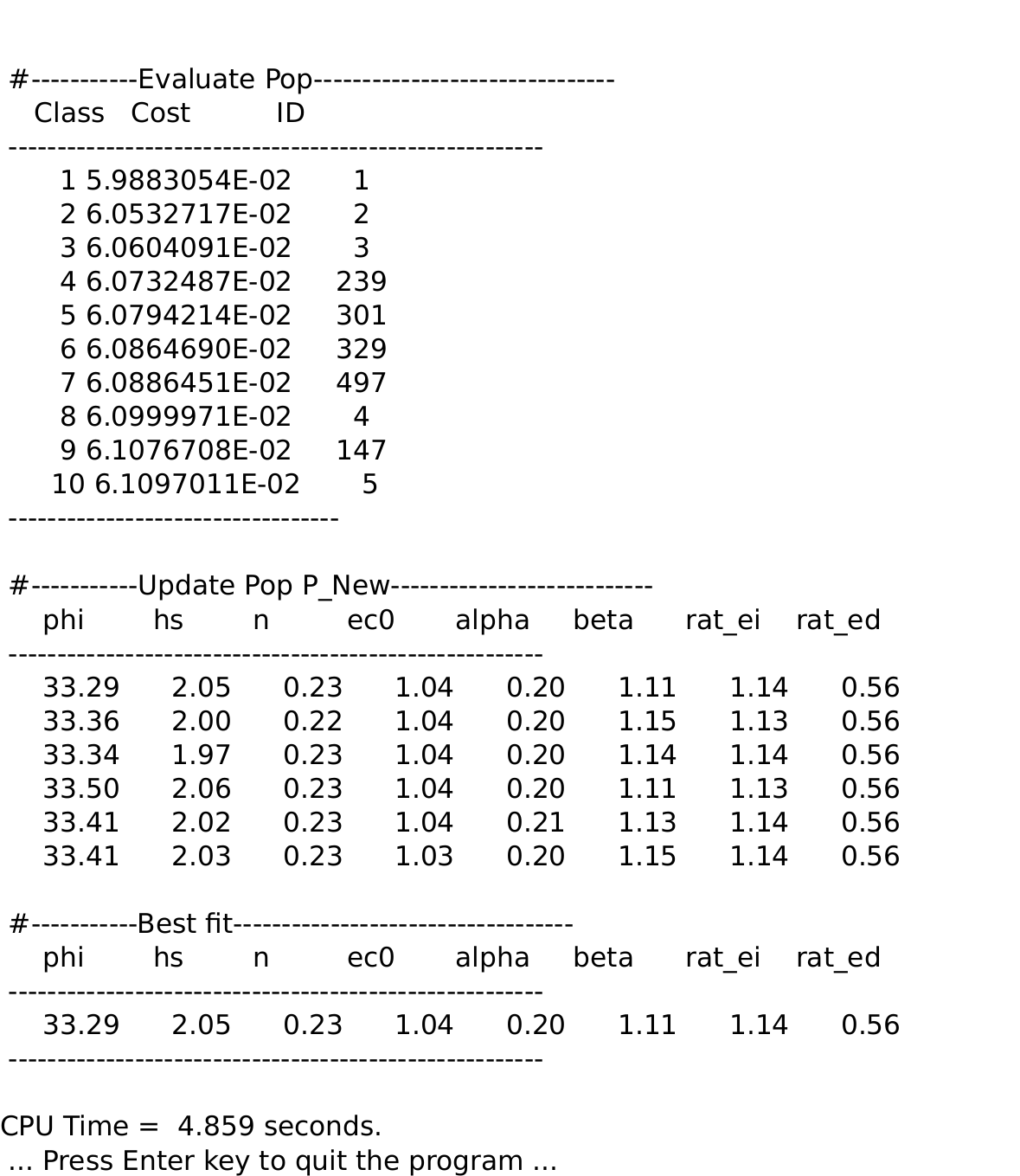}}}
  \caption{The command-line interface of the GA-cal in three different stages. (a) Input part with the highlighting box of the input file specification,  (b) convergence check, (c) final tables output.}
\label{fig:CLI}
\end{figure}

Launching the executable produces the command line interface in (CLI) in Figure \ref{fig:CLI_1}, where the input file should be specified. Once again, all input files should be in a subfolder \textit{data} on the same path from which the executable is launched.

During execution, the program writes a series of information to the CLI which can be divided into three stages:

\begin{enumerate}
    \item \textbf{summary of input files}. The first lines of the CLI show all the data related to the program input files described in the previous section. The output on the CLI is here not shown for brevity.

    \item \textbf{the evolution of the GA optimization} (see Figure \ref{fig:CLI_2}). Starting from the initial population, the first line displays the iteration counter while the others report some relevant information concerning the model performances in the three previously described planes. For each of these, the table reports the maximum and minimum values of the deviations $ \delta_i$ (see eq. \eqref{eq:delta_i}), that is delta OE, delta TD\_q and delta TD\_e. The best score is also printed; this is the minimum value of the cost function (see eq. \ref{eq:Sco}). These outputs allow for evaluating the optimizer's convergence during the iterations. In general, if one seeks a calibration that does not favour any particular plane, it suffices using weights of comparable orders of magnitude.
    
    The user can control the weights $w_i $ in the eq. \eqref{eq:Sco}, by editing the \textit{Input.txt} files, if the error in one particular plane is overly large or if a specific plane needs to be prioritized. This might occur, for example, in the stability analysis of a slope, for which the behaviour in the plane ($\varepsilon_a,q$) is far more important than the behaviour in the plane ($\sigma_v,e$);
    
    \item \textbf{the summary from the optimization} (see Figure \ref{fig:CLI_3}). This includes three tables. The first table reports on the performances of the best ten candidate solutions and their associated cost function and ID in the last iteration. The candidate ID is simply the row number of the $\mathbf{Pop}$ matrix.
    
    Note that the ID is assigned after each population update, while the ranking is carried out after each population evaluation. In the specific example shown in Figure \ref{fig:CLI_3}, in which five elites are chosen at each iteration, four candidates (IDs 239, 301, 239, 497) have performed better than the fourth elite. Therefore, their ID would change in the following population update. However, the algorithm closes the last iteration with an evaluation (not an update) hence the mismatch between ranking and IDs, which helps understand the degree to which the best candidates have converged. 
    
    The second table illustrate the parameters of the best five candidates (in this case IDs 1, 2, 3, 239, 301). The closeness of these parameters proves the convergence of the optimizer, as the population cannot settle over different local minima if a sufficiently large number of iterations is performed unless these minima have the same (to machine precision) value of the cost function. The occurrence of such a highly unlike event can nevertheless be spotted by analyzing this table together with the first.
    
    In case of large variation of the parameters, the user should thus change the setting of the optimizer in the \textit{GA\_setup.txt} (increasing the number of iterations, number of individuals, and /or decreasing the mutation factors). Finally, the last table shows the values of the parameter optimized (Best fit), while the last lines show the calculation time.
    
\end{enumerate}

The results from the tutorial calibration are reported in Table \ref{tab:W_H_GA_confronto}. Being the GA a stochastic optimizer, these values might differ at each run; however, the results for this problem should not differ much and are indeed similar to those reported in the literature for this test case. In an ongoing work by the authors \cite{TOAPPEAR}, the intrinsic variability of the GA is used to analyze the uncertainty in the model prediction (see also \cite{10.1007/978-3-031-12851-6_32}) and the problem of parameter uniqueness. 

\begin{figure}[htbp]
  \centering
  \subcaptionbox{Oedometric plane}[1\linewidth][c]{%
    \includegraphics[width=0.44\textwidth]{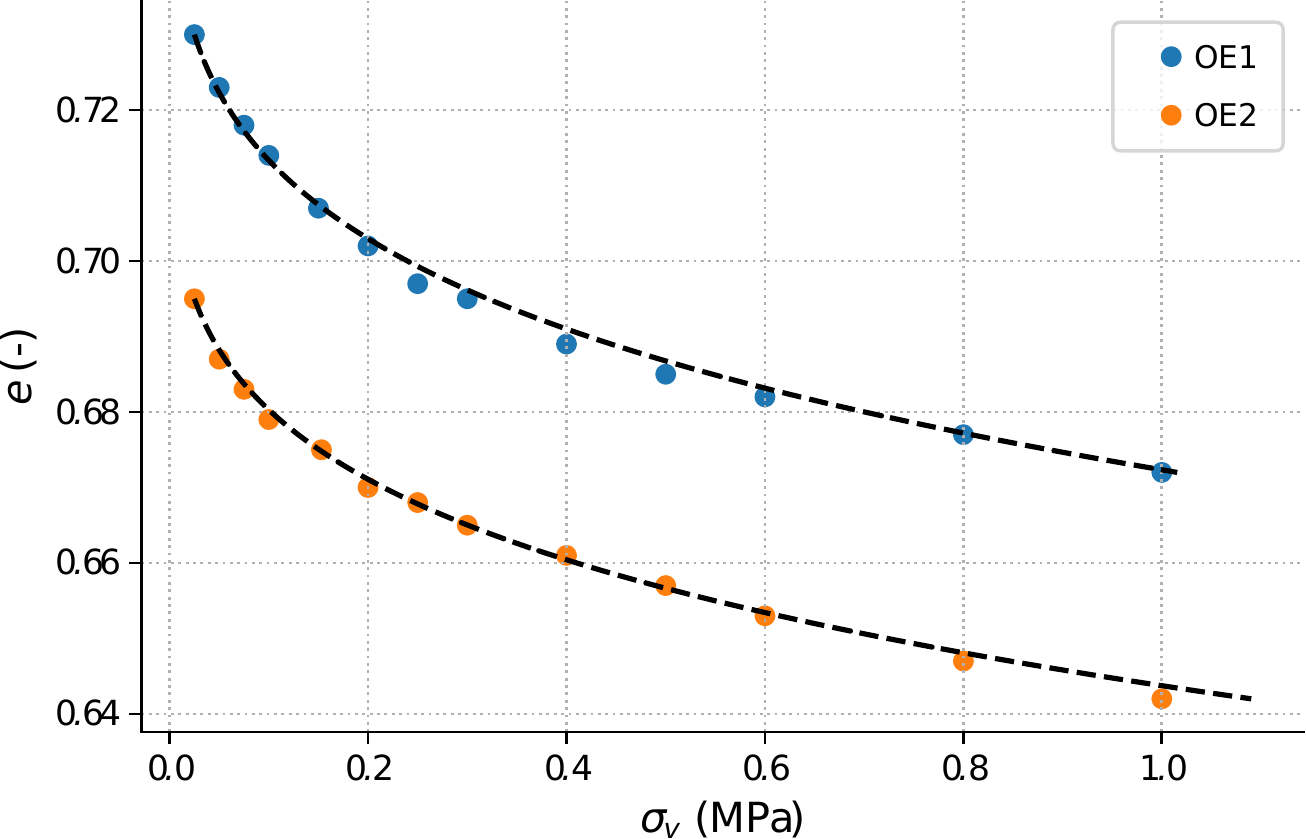}}\\
  \subcaptionbox{Triaxial deviatoric plane}[1\linewidth][c]{%
    \includegraphics[width=0.44\textwidth]{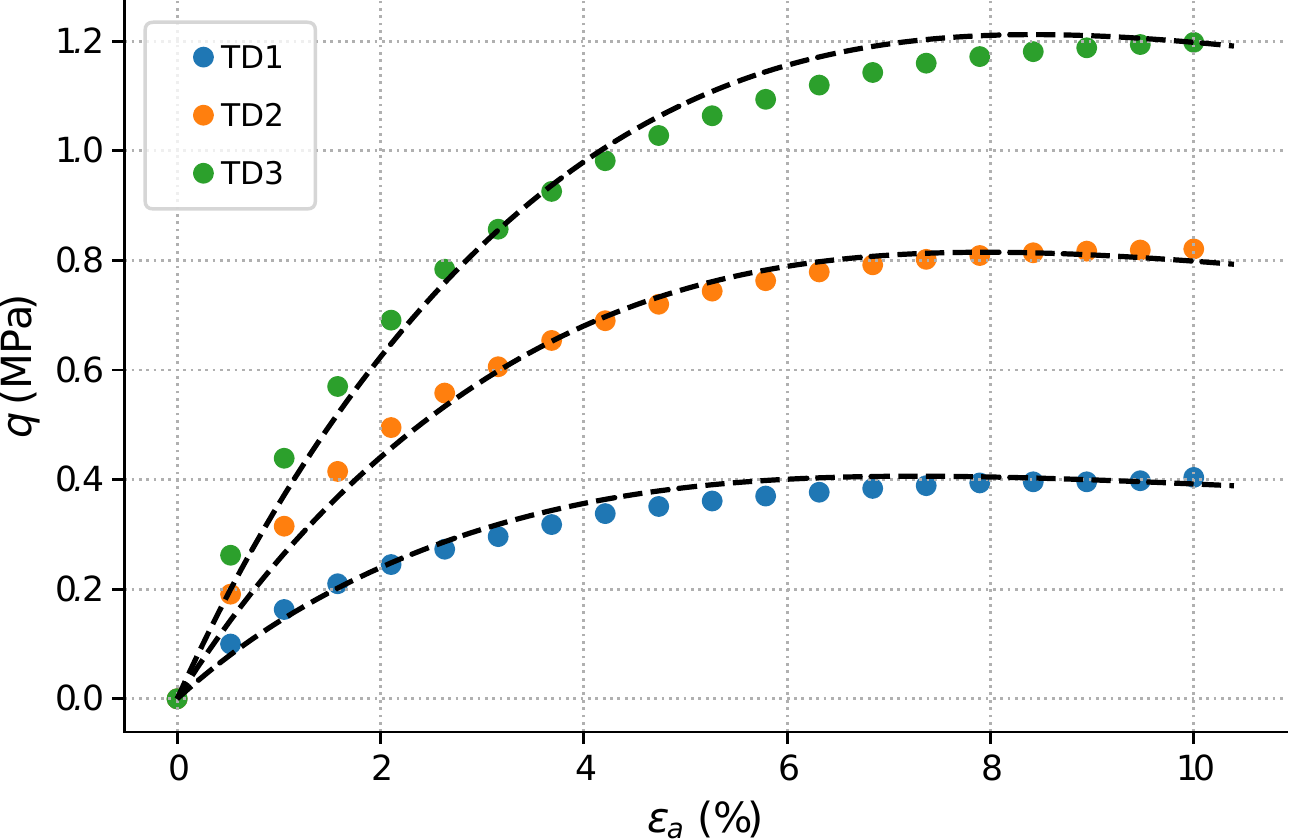}}\\
  \subcaptionbox{Triaxial volumetric plane}[1\linewidth][c]{%
    \includegraphics[width=0.44\textwidth]{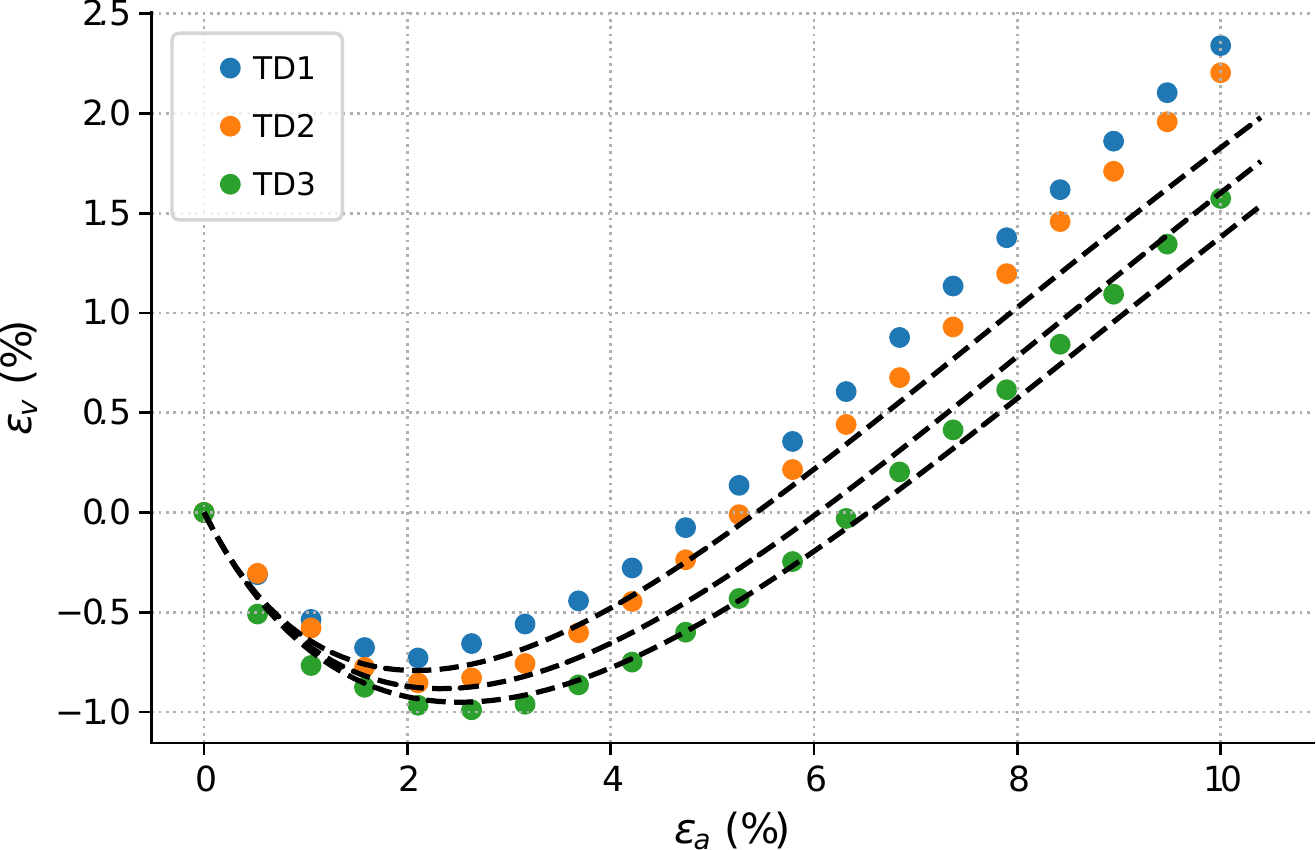}}
  \caption{Comparison between the response curves of the SH model calibrated with GA-cal and the experimental data (point) on Hochstetten sand \cite{Wolffersdorff_A_hypoplastic_for_granular_material_with_a_predefined_limit_state_surface}. }
\label{fig:GA-cal_curve}
\end{figure}

The plots in Figure \ref{fig:GA-cal_curve}, obtained with the script \textit{Py\_Out.py} attached to the code documentation, show the excellent agreement of the calibrated SH model with the experimental data.
In terms of cost function and the response curve of the model, all the combinations suggested by GA-cal with different executions are practically identical when the optimization parameters allow convergence. Moreover, as reported in \cite{Mendez2021}, correlations can emerge from the parameters obtained from different calibrations.


\begin{table}[h]
 \centering
  \caption{SH model parameter for the Hochstetten sand from von Wolffersdorff (W) \cite{Wolffersdorff_A_hypoplastic_for_granular_material_with_a_predefined_limit_state_surface}, Herel and Gudehus (H) \cite{Herle_Gudehus_Determination_of_parameters_of_a_hypoplastic_constitutive_model_from_properties_of_grain_assemblies},  Mendez et al. (M) \cite{Mendez2021} and  GA-cal  (GA-cal).}
    \begin{tabular}{lcrrrr}
    \toprule 
    \multicolumn{2}{l}{Par.} & \multicolumn{1}{c}{W} & \multicolumn{1}{c}{H} &  \multicolumn{1}{c}{M} & \multicolumn{1}{c}{GA-cal}\\
    \midrule
    $\varphi$&($^{\circ}$) & 33.00 &  33.00  & 32.73 & 33.52 \\
    $h_s$    &($GPa$)      &  1.00 &  1.50   & 1.32  &  2.22 \\
    $n$      &($-$)        &  0.25 &  0.28   & 0.23  &  0.22 \\
    $e_{i0}$ &($-$)        &  0.55 &  0.55   & 0.60  &  0.58 \\
    $e_{c0}$ &($-$)        &  0.95 &  0.95   & 1.04  &  1.03 \\
    $e_{d0}$ &($-$)        &  1.05 &  1.05   & 1.14  &  1.16 \\
    $\alpha$ &($-$)        &  0.25 &  0.25   & 0.23  &  0.21 \\
    $\beta$  &($-$)        &  1.50 &  1.00   & 1.26  &  1.12 \\
    \bottomrule
    \end{tabular}%
  \label{tab:W_H_GA_confronto}%
\end{table}%

\subsection*{Acknowledgements}
The authors gratefully acknowledge the support and the discussions with the engineer Pierantonio Cascioli, from GEINA srl, and Gabriele Sandro Toro, laboratory technician of the Department of Engineering and Geology of the Faculty Gabriele D'Annunzio of Chieti.

{\small
\bibliographystyle{ieee_fullname}
\bibliography{Bibliography}
}


\end{document}